\documentclass[conference]{IEEEtran}
\IEEEoverridecommandlockouts
\usepackage{cite}
\usepackage{amsmath,amssymb,amsfonts}
\usepackage{algorithmic}
\usepackage{graphicx}
\usepackage{textcomp}
\usepackage{xcolor}
\usepackage[hyphens]{url}
\usepackage{balance}
\usepackage{graphicx}

\usepackage{adjustbox}
\usepackage{afterpage}
\usepackage[ruled,vlined]{algorithm2e}
\usepackage{tikz}
\usetikzlibrary{shapes,arrows,calc,positioning}
\usetikzlibrary{automata}
\usepackage{amsmath,amssymb,mathtools}
\usepackage{xcolor}
\usepackage{tabularx,booktabs}
\usepackage{url}
\usepackage{breakurl}
\usepackage{wrapfig}
\usetikzlibrary{arrows.meta}
\usepackage{subcaption}
\usepackage{psfrag}
\usepackage{epstopdf}
\usepackage[justification=centering]{caption}
\usepackage{pgfplots}
\usepackage{gensymb}
\usepackage{psfrag}
\usepackage{epstopdf}
\usepackage{tablefootnote}
\usepackage{caption}
\usepackage{ifthen}
\usepackage{cleveref}
\pgfplotsset{compat=newest}
\usetikzlibrary{plotmarks}
\usepackage{multirow}

\def\BibTeX{{\rm B\kern-.05em{\sc i\kern-.025em b}\kern-.08em
    T\kern-.1667em\lower.7ex\hbox{E}\kern-.125emX}}
\begin{document}
\setlength{\abovedisplayskip}{0pt}
\setlength{\belowdisplayskip}{0pt}

\title{False Data Injection Attacks in Internet of Things and Deep Learning enabled Predictive Analytics\\
{\footnotesize}
\thanks{This is the extended version of the paper entitled ``\emph{Impact of False Data Injection Attacks on Deep Learning enabled Predictive Analytics}", accepted for publication in the 32nd IEEE/IFIP Network Operations and Management Symposium (NOMS 2020).}
}

\author{\IEEEauthorblockN{Gautam Raj Mode, Prasad Calyam, Khaza Anuarul Hoque}
\IEEEauthorblockA{\textit{Department of Electrical Engineering \& Computer Science} \\
\textit{University of Missouri, Columbia, MO, USA}\\
gmwyc@mail.missouri.edu, calyamp@missouri.edu, hoquek@missouri.edu}
}

\maketitle

\begin{abstract}
Industry 4.0 is the latest industrial revolution primarily merging automation with advanced manufacturing to reduce direct human effort and resources. Predictive maintenance (PdM) is an industry 4.0 solution, which facilitates predicting faults in a component or a system powered by state-of-the-art machine learning (ML) algorithms (especially deep learning algorithms) and the Internet-of-Things (IoT) sensors. However, IoT sensors and deep learning (DL) algorithms, both are known for their vulnerabilities to cyber-attacks. In the context of PdM systems, such attacks can have catastrophic consequences as they are hard to detect due to the nature of the attack. To date, the majority of the published literature focuses on the accuracy of DL enabled PdM systems and often ignores the effect of such attacks. In this paper, we demonstrate the effect of IoT sensor attacks (in the form of false data injection attack) on a PdM system. At first, we use three state-of-the-art DL algorithms, specifically, Long Short-Term Memory (LSTM), Gated Recurrent Unit (GRU), and Convolutional Neural Network (CNN) for predicting the Remaining Useful Life (RUL) of a turbofan engine using NASA's C-MAPSS dataset. The obtained results show that the GRU-based PdM model outperforms some of the recent literature on RUL prediction using the C-MAPSS dataset. Afterward, we model and apply two different types of false data injection attacks (FDIA), specifically, continuous and interim FDIAs on turbofan engine sensor data and evaluate their impact on CNN, LSTM, and GRU-based PdM systems. The obtained results demonstrate that FDI attacks on even a few IoT sensors can strongly defect the RUL prediction in all cases. However, the GRU-based PdM model performs better in terms of accuracy and resiliency to FDIA. Lastly, we perform a study on the GRU-based PdM model using four different GRU networks with different sequence lengths. Our experiments reveal an interesting relationship between the accuracy, resiliency and sequence length for the GRU-based PdM models.
\end{abstract}


\section{Introduction}
Current advances in machine learning (ML) techniques and Internet-of-Things (IoT) sensors has enabled the emergence of predictive maintenance (PdM), which is a method of preventing asset failure by analyzing production data and identifying patterns to predict issues before they happen. State-of-the-art PdM techniques can help reduce downtime by 35\%-45\%, maintenance cost by 20\%-25\%, and can increase production by 20\%-25\%~\cite{ibm}. Due to these benefits, IoT and ML-enabled PdM solutions are reshaping automotive, aerospace, oil and gas, transportation, manufacturing industries and also reshaping the national defense. Specifically, deep learning (DL) algorithms have recently shown tremendous success in such PdM applications~\cite{der2019applying}.  Unfortunately, IoT sensors and DL algorithms are both susceptible to attacks~\cite{sikder2018survey}, which poses a significant threat to the overall PdM system. According to a recent report from the \emph{Malwarebytes}, cyber-threats against businesses/factories have increased by more than 200\% over the past year~\cite{malware}.

Specifically, it is very hard to detect stealthy attacks, such as False Data Injection Attack (FDIA)~\cite{FDIA} on the PdM system due to the nature of the attack. In false data injection attack (FDIA) \cite{FDIA}, an attacker stealthily compromises measurements from IoT sensors (by a very small margin), such that the manipulated sensor measurements bypass the sensor's basic `faulty data' detection mechanism and propagates to the sensor output undetected. An FDI attack can be implemented by compromising physical sensors, sensor communication network, and data processing programs. Such attacks on a PdM system may not even show their impact. Instead, the attack propagates from the sensor to the ML part of the PdM system and fools the system by predicting a delayed asset failure or maintenance interval. This might incur a significant cost by inducing an unplanned failure or loss of human lives in safety-critical applications~\cite{aero11,auto22,rail}. 

FDI attacks have already caused many known disastrous incidents, such as the Northeast blackout of 2003 in the USA and the Ukrainian power grid attack affecting over 230,000 people, leaving them without electricity for several hours. Extensive research has been performed on the detection and mitigation of FDI attacks in cyber-physical systems (CPS) domain \cite{CPS1,CPS2,CPS3}. Unfortunately, the effect of FDIA on a PdM system is yet not explored which motivates our research. In the case of aircraft engine PdM systems, FDIAs may result in the delay of timely maintenance and lead to mid-air engine failures which are catastrophic. Current users of PdM systems for aircraft engine maintenance include Pratt and Whitney, Rolls-Royce, Honeywell, General electronics and the US Air force~\cite{aero11,aero22,aero33,aero44}. For example, Bombardier’s new jetliner uses a Pratt and Whitney turbofan engine that boasted more than 5,000 sensors~\cite{aero55, aero66}. Powered with the modern DL algorithms, this engine can predict the future demands of the engine, perform adjustments, and thus save 15\% of fuel usage. However, the vulnerability of sensor-attacks against for such IoT and ML-based engines is considered a challenge~\cite{aero55, mot1, mot3}. The existing sensor attack detection solutions in the IoT and cyber-physical system domain is not sufficient to address this problem due to the fact that, when deployed individually to the thousands of sensors, most of the existing techniques suffer from scalability problems and resource overheads as many IoT sensors are power and resource-constrained.\\

\begin{figure}[t]
	\centering
	\includegraphics[width=0.45\textwidth]{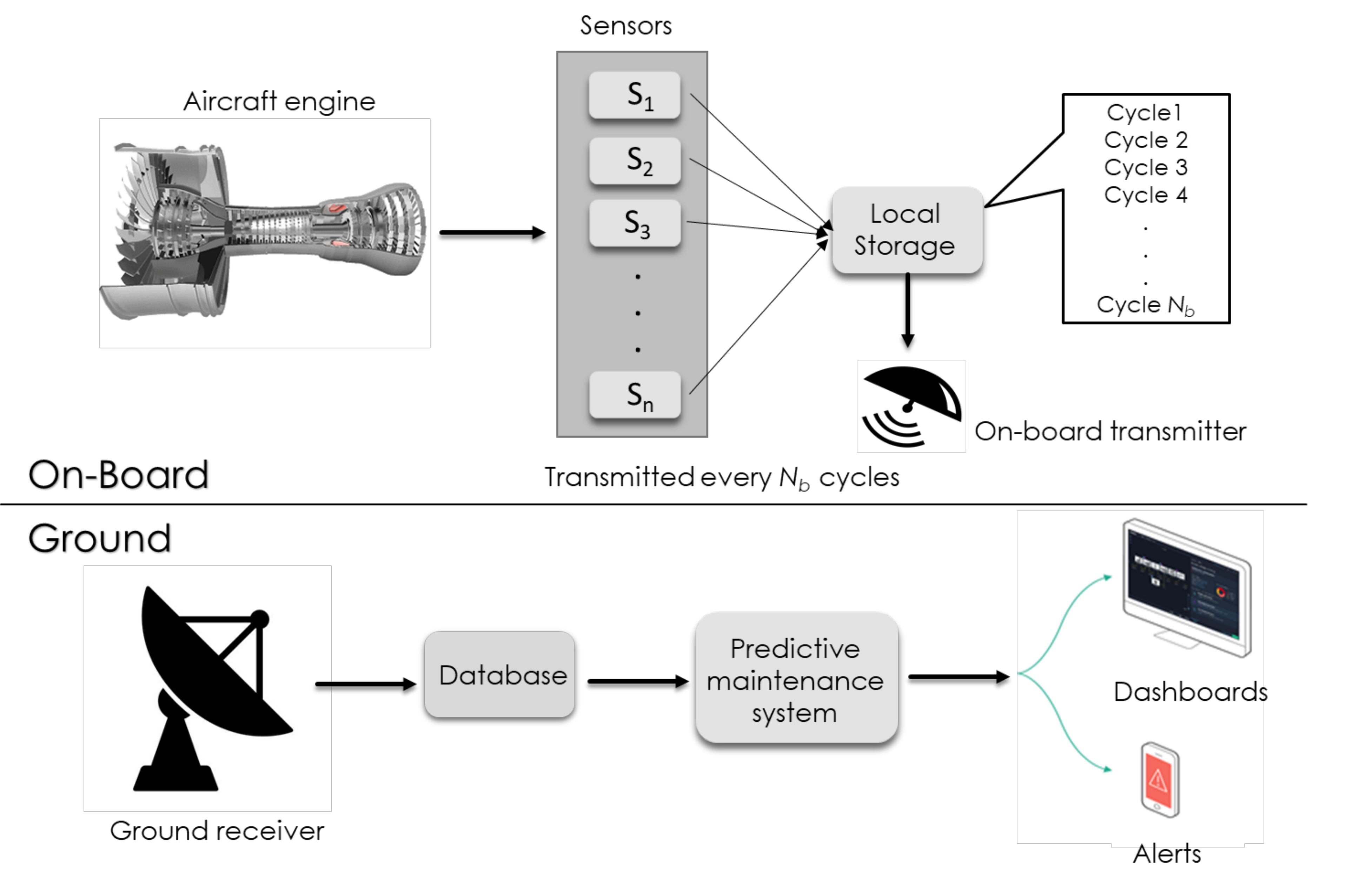}
	\caption{Engine health monitoring (EHM) system architecture}
	\label{fig:EHM}
\end{figure}

\noindent \emph{Contribution of this paper}: In this paper, we model continuous and interim FDIAs on IoT sensors and show their impact on a PdM model by performing a case study on the aircraft Predictive Maintenance (PdM) system. We use the  C-MAPSS~\cite{saxena2008c} (Commercial Modular Aero-Propulsion System Simulation) dataset\footnote{a popular turbofan engine degradation dataset published by NASA's Prognostics Center of Excellence (PCoE)}. At first, to build an accurate predictive model, we train the Long Short-Term Memory (LSTM), Gated recurrent unit (GRU), and Convolutional neural network (CNN) algorithms using the C-MAPSS dataset. We evaluate these three predictive models, and the obtained results show that the GRU-based model predicts the Remaining Useful Life (RUL)\footnote{Remaining useful life (RUL) is the length of time a machine is likely to operate before it requires repair or replacement.} most accurately. The obtained results from the GRU-based model outperforms the recent works that use DL for RUL prediction using the C-MAPSS dataset in~\cite{zheng2018data, zheng2017long,ellefsen2019remaining} (by predicting RUL 1.3-1.9 times more accurately). 

Afterward, we model two types of false data injection attacks (FDIA) on the C-MAPSS dataset and evaluate their impact on CNN, LSTM, and GRU-based PdM models. To be more realistic, we model attack only on 3 sensors among the 21 sensors in the turbofan engine. The obtained results show that all the PdM models are greatly defected by the FDIA even if only 3 out of the 21 sensors are attacked. However, the GRU-based PdM model is comparatively more accurate and resilient to FDIA when compared to the other evaluated PdM models. In terms of sensitivity, we also explore that CNN is way more sensitive to FDIAs when compared to the LSTM and GRU. This is indeed an important observation since CNN-based techniques are quite popular in asset maintenance~\cite{silva2019cnn,huuhtanen2018predictive,caponetto2019deep} and our results indicate that special measures should be taken for designing a CNN-based PdM. Afterward, we analyze the GRU-based PdM model using four different sequence lengths. The obtained results show an interesting relationship between the accuracy, the resiliency and the sequence length of the models. To the best of our knowledge, this is the \emph{first work} that demonstrates the effects of IoT sensor attacks on a deep learning-enabled PdM system. \\

\noindent \emph{Paper organization}: The rest of the paper is organized as follows. Section II briefly discusses the Engine Health Monitoring (EHM) systems and Predictive Maintenance (PdM). Section III introduces the DL algorithms used in this paper: LSTM, GRU, and CNN, and also describes NASA's C-MAPSS dataset used for our experiment. Section IV describes the modeling of FDIA in detail. Section V compares the performance of CNN, LSTM, and GRU in predicting RUL and analyses the impact on RUL prediction using CNN, LSTM, and GRU after both continuous and interim FDI. Section VI presents the observations from those obtained results, and Section VII concludes the paper.




\section{Engine health monitoring (EHM) system}

\subsection{Engine health monitoring (EHM) system}
An aircraft engine is a complex system, so it requires adequate monitoring to ensure safe operation and in-time maintenance ~\cite{airplane}. Several displays and dials in the cockpit give different measurements like exhaust gas temperatures, engine pressure ratio, the pressure at fan inlet, rotational speeds, etc. All these parameters are crucial in indicating the health of the engine; they serve as early indicators of failure and prevent costly component damage. In order to accomplish the task of monitoring these parameters in an engine, Engine health monitoring (EHM) systems ~\cite{EHM} have been in service for three decades. Fig.  \ref{fig:EHM} shows a generic EHM architecture. An EHM system has several IoT (Internet of Things) sensors mounted inside and outside of an engine to monitor different parameters. All these IoT sensors are connected to a wireless network ~\cite{wirelessSensors}, which uses radiofrequency for transmitting sensor output to central engine control ~\cite{wirelessSensors}. These IoT sensors monitor different parameters of an aircraft engine and send out alerts to the engine manufacturer if the RUL \cite{RUL} of the engine is approaching its end of life. An EHM system employs PdM systems to predict the RUL using the data collected from the IoT sensors.

The sensors on-board the engine send time-series data (cycles) every hour to the local storage on-board the airplane. After every $N_b$ cycles of data are captured, the data is transmitted to the ground station. At the ground station, the incoming live data is stored in the database and sent to the PdM system to predict RUL of the engine. The PdM system sends out alerts if the predicted RUL is less than the permissible safe operation RUL of the engine.

\subsection{Predictive maintenance (PdM)}
In manufacturing supply chains, unexpected failures are considered as primary operational risk as it can hinder productivity and can incur huge losses. For example, in the modern automotive industry, an assembly line has several robots working on a car, and even if one of the robots fails, it will result in the total halt of the assembly line, causing loss of valuable production time and money. To overcome this problem, PdM strategies are employed. PdM is an industry 4.0 solution, which assists in predicting the future state of physical assets. It helps in better-informed maintenance decisions, to prevent unexpected delays.


PdM systems are employed in major industries like Nuclear power plants, aviation industry, automotive industry, and health care services. PdM allows for convenient scheduling of corrective maintenance as parts for the equipment can be ordered beforehand to avoid the last-minute hassle, which saves a lot of valuable production time. PdM is well suited for making an informed decision when dealing with time-series data. A data-driven model of PdM employs some of the remarkable strategies like the Random Forest algorithm \cite{RFA}, Artificial Neural Networks (ANN) \cite{ANN}, fuzzy models \cite{fuzzy}, Big data frameworks \cite{BigData}. In this paper, three deep learning algorithms, specifically, LSTM, GRU, and CNN are employed in predicting RUL of an aircraft engine.

 




\section{DL algorithms for RUL prediction}


As mentioned earlier, RUL can be predicted using different ML algorithms. For this paper, we utilize LSTM, GRU, and CNN algorithms and compare their performance. 

\subsection{Long short-term memory model (LSTM)}
An LSTM \cite{Back2} is a special kind of Recursive Neural Network (RNN), capable of learning long-term dependencies. LSTM is explicitly designed to avoid long term dependency problems, which is prevalent in RNN. It has achieved great praise in the field of machine learning and speech recognition. Some of the neural networks have a dependency problem, but an LSTM can overcome the problem of dependency by controlling the flow of information using input, output and forget gate. The input gate controls the flow of input activation into the memory cell. The output gate controls the output flow of cell activation into the rest of the network.


Suppose that training data has $N$ equipment of the same make and type that provide failure data, and each equipment provides set multivariate time-series data from the sensors of the equipment.  Also, assume that there are $r$ sensors of the same type on each equipment. Then data collected from each equipment can be represented in a matrix form $X_n = [x_1, x_2, ..., x_t, ..., x_{T_n} ] \in \mathbb{R}^{r \times T_n} ~(n = 1, ..., N )$ where $T_n$  is  time of  the  failure and at time $t$ the $r$-dimensional vector of sensor measurements is $x_t  = [s_t^1 , ..., s_t^r ] \in \mathbb{R}^{r \times 1}, t = 1, 2, ..., T_n$. 
The data of each equipment in $X_n$ is fed to LSTM network and the network learns how to model the whole sequence with respect to target RUL. At time $t$, LSTM network takes $r$-dimensional sensor data $x_t$ and gives predicted $RUL_t$.


Let the LSTM cell has $q$ nodes, then $c_t \in \mathbb{R}^{q \times 1}$ is output of cell state, $h_t \in \mathbb{R}^{q \times 1}$ is output of LSTM cell, $o_t \in \mathbb{R}^{q \times 1}$ is output gate, $i_t \in \mathbb{R}^{q \times 1}$ is input gate, and $f_t \in \mathbb{R}^{q \times 1}$ is forget gate at time $t$. At time $t-1$, the output $h_{t-1}$, and hidden state $c_{t-1}$ will serve as input to LSTM cell at time $t$. The input $x_t$ is fed as input to the cell. In LSTM, the normalized data are calculated using the following equations:
\begin{align} \label{eq:normalization}
i_t = \sigma(W_i\cdot [h_{t-1},x_t]+b_i), \\ 
f_t = \sigma(W_f\cdot [h_{t-1},x_t]+b_f),\\
o_t = \sigma(W_o\cdot [h_{t-1},x_t]+b_o),\\
\widetilde{c}_t = act(W_c\cdot [h_{t-1},x_t]+b_c),\\
c_t = f_t \ast c_{t-1} + i_t \ast \widetilde{c}_t,\\
h_t = o_t \ast act(c_t),
\end{align}

Where $\sigma$ is the sigmoid layer. $c_t$ and $\widetilde{c}_t$ are each internal memory cell and temporary value to make a new internal memory cell at time t. $\ast$ is element-wise multiplication of two vectors.

\subsection{Gated recurrent unit (GRU)}
The GRU was proposed by \emph{Cho et al.} \cite{cho2014properties}. It operates using a reset gate and an update gate. The GRU is an improved version of standard recurrent neural networks. Similar to the LSTM unit, the GRU has gating units that modulate the flow of information, however, without having a separate memory cell. GRU's performance on certain tasks of polyphonic music modeling and speech signal modeling was found to be similar to that of LSTM. GRUs have been shown to exhibit even better performance on certain smaller datasets.\cite{zhao2017machine}. 
The memory block of GRU is simpler than that of LSTM. The forget, input and output gates are replaced with an update and a reset gate. Also, GRU combines the hidden state and the internal memory cell. In GRU, the normalized data are calculated using the following equations:
\begin{align} 
z_t = \sigma(W_z\cdot [h_{t-1},x_t]+b_z), \\ 
r_t = \sigma(W_r\cdot [h_{t-1},x_t]+b_r),\\
\widetilde{h}_t = act(W \cdot [r_t \ast h_{t-1},x_t]+b_h),\\
h_t = (1-z_t) \ast h_{t-1} + z_t \ast \widetilde{h}_t,
\end{align}

where $z_t$ and $r_t$ are the update gate and reset gate at time $t$, respectively. $\widetilde{h}_t$ is a temporary value to make new hidden state at time $t$.

\subsection{Convolutional neural network (CNN)} 

CNN a deep learning algorithm has achieved exceptional success in various research fields \cite{bhandare2016applications} because it has many advantages over traditional machine learning approaches such as MLP \cite{swain2016fuzzy}. CNNs are fundamentally inspired from feed-forward ANNs. Like any other advanced DL algorithm, CNN also finds its applications in different areas including CNN-based PdM system~\cite{silva2019cnn,huuhtanen2018predictive,caponetto2019deep}. A CNN consists of one or more convolutional layers and then followed by one or more fully connected layers as in a standard multi-layer neural network. A 1D CNN model is utilized in this paper to predict the RUL of the engine. Details about CNN construction and network design are presented in detail in \cite{ince2016real}.


\subsection{C-MAPSS dataset}
To evaluate the performance of the CNN, LSTM, and GRU DL algorithms, we use a well-known dataset, NASA's turbofan engine degradation simulation dataset C-MAPSS (Commercial Modular Aero-Propulsion System Simulation). This dataset includes 21 sensor data with different number of operating conditions and fault conditions\footnote{More details about these 21 sensors can be found in \cite{Dataset} and \cite{sensorsTable}}. In this dataset, there are four sub-datasets (FD001-04). Every subset has training data and test data. The test data has run to failure data from several engines of the same type. Each row in test data is a time cycle which can be defined as an hour of operation. A time cycle has 26 columns where the 1st column represents the engine ID, and the 2nd column represents the current operational cycle number. The columns from 3 to 5 represent the three operational settings and columns from 6-26 represent the 21 sensor values. The time-series data terminates only when a fault is encountered. For example, an engine with ID 1 has 192 time cycles of data, which means the engine has developed a fault at the 192nd time cycle. The test data contains data only for some time cycles as our goal is to estimate the remaining operational time cycles before a fault.

\section{Modeling of FDIA} \label{Sec:attacks}
In this section, we describe the modeling of FDIAs, attacker's objective, and the attack scenarios in detail.\\

\noindent \textbf{False data injection attack (FDIA):} As mentioned earlier, false data injection attack (FDIA) \cite{FDIA} can be injected into the system by compromising physical sensors, sensor data communication links, and data processing programs. Compromising physical sensors requires physical access to the sensors and hence is a tedious task. In contrast, hacking the sensor data communication links and data processing programs is an easier option for an attacker (explained in detail in the \emph{attack surface} of this section). A successful FDIA can cause the engine sensors to output erroneous values to the central engine control, and thus make either physical or economic impact on the predictive maintenance model. For example, $X_{i}$ represents the information transmitted by the $i^{th}$  sensor. In an FDIA, the adversary contaminates the original vector with a vicious vector. Let $X_{i} = [x_{1},x_{2},...,x_{k}]$ be the original vector data containing $k$ sensor reading for the $i^{th}$ sensor. The original vector could be contaminated by adding an FDIA vector with the same dimension as the original vector. Let the contaminated vector for the $i^{th}$ sensor be $F_{i}=[\lambda_1,\lambda_2,...,\lambda_k]$, then the compromised vector is given by Eq. \ref{eq:FDIA}.

\begin{equation} \label{eq:FDIA}
Z_{i}=X_{i}+F_{i}
\end{equation}


An FDIA can be \emph{constrained}, where the attacker has access to a limited number of sensors, and some part of the communication network and an FDIA can also be \emph{unconstrained}, where the attacker has access to all of the sensors and also has total control of the communication network. In this work, we consider the constrained attack since it is more practical that an attacker has access to only a limited number of sensors (for the case study, the attack scenario considers only 3 sensors from a total of 21 sensors). We model two variations of FDIAs to explore and compare their impact, specifically,  \emph{continuous FDIA} and \emph{interim FDIA}. In the case of continuous FDIA, the attack is continuous, which means, once the attack starts, from that point onwards all the sensor readings are compromised. For instance, if the attack starts at the time instant $atck\_start=3$ and ends at $atck\_end$ then $F_i$ can be expressed as $F_{i}=[\lambda_1,\lambda_2, \lambda_{atck\_start},...,\lambda_{atck\_end}]$, where $atck\_start\geq1$ and ${atck\_end}=k$ . In the case of interim FDIA, the duration of attack is a short time interval, where $atck\_start > 1$ and  ${atck\_end} < k$.\\

\noindent \textbf{{Attacker's stealthiness:}} An FDIA can be stealthy if it is not detected by the defense mechanism. In order to achieve that objective, the attack vector should remain in the boundary conditions of the sensor measurements. There exist constant vectors $Z_{min}$ and $Z_{max}$, such that for any FDIA vector $Z_i$, the compromised vector passes undetected through the defense if

\begin{equation} \label{eq:FDIAStealth}
Z_{i}=X_{i}+F_{i}~ and~ Z_{min} \leq Z_i \leq Z_{max}
\end{equation}
We assume the attacker knows $Z_{min}$ and $Z_{max}$ to construct attack vectors satisfying Eq.\ref{eq:FDIAStealth}. Such information is easily available from the sensor data sheets provided by the vendor. \\

\noindent \textbf{Attacker's objective:}
The attacker's objective is to cause a delay in aircraft engine maintenance. This objective can be achieved by altering the IoT sensors readings that are fed to the PdM systems. Injecting false data to the sensor readings result in incorrect predictions from PdM systems which in turn results in a delay of timely maintenance. As timely maintenance is a crucial factor in engine performance, a lapse of maintenance may result in mid-air engine failures which are catastrophic. One can argue that the attacker having access to the physical sensors or the communication network of the sensors would directly attack the main systems (flight navigation and instrument landing systems) rather than just altering the sensor values for the PdM. However, there is a higher chance that a direct attack on the main system will easily get detected by the defense mechanisms. In contrast, introducing FDIA to sensors is a safer option for an attacker since such attacks are more stealthy, hard to detect as they are in the sensor's acceptable range. Thus, such attacks will cause an erroneous calculation of the RUL and might delay the maintenance cycle leading to a catastrophic incident.\\


\noindent \textbf{Attack surface:}
In this paper, only the \emph{constrained attacks} are considered. Note, one of the ways to launch an FDIA is using spoofing techniques. For instance, Tippenhauer \emph{et al.} \cite{tippenhauer2011requirements} showed a spoof attack scenario on GPS-enabled devices. In this attack scenario, a forged GPS signal is transmitted to the device to alter the location. In this way, the true location of the device is disguised and the attacker can perform a physical attack on the device. In another work, Giannetos \emph{et al.} \cite{giannetsos2013spy} introduced an app named \emph{Spy-sense}, which monitors behaviors of several sensors in a device. The app can manipulate sensor data by deleting or modifying it. \emph{Spy-sense} exploits the active memory region in a device and relays sensitive data covertly. These works show that FDI attacks can be performed even without gaining direct access to a system. 

One of the recent articles~\cite{mcas} considers cyber-attacks as one of the reasons behind the two recent Boeing 737 Max 8 crashes. According to that article, a passenger, vehicle or drone carrying a sonic device capable of impacting the MCAS sensor controlling the plane could have been responsible for such an attack. Recently, ICS-CERT published an alert on certain controlled area network (CAN) bus systems aboard aircraft that might be vulnerable to hacking. It cited a report that an attacker with access to the aircraft could attach a device to avionics CAN bus to inject false data, resulting in incorrect readings in an avionic equipment \cite{icsAlert}. Using such a device attached to the bus could lead to incorrect engine telemetry readings, compass, altitude, airspeed, and angle of attack (AoA) data. Pilots might not be able to distinguish between false and legitimate readings. This alert explores the possibility of injecting false data into IoT sensor readings of aircraft engine which are transmitted on a CAN. In this work we consider FDIA using a malicious device attached to avionics CAN.\\

\noindent \textbf{Attack scenario:}
As shown in Fig. \ref{fig:EHM} of the EHM architecture, the aircraft sends $N_b$ cycles of data at a time to the ground station/engine manufacturer. At the ground station, the PdM system performs data analytics on the received data and send out alerts if the RUL is close to the threshold $N_{th}$. The value of $N_{th}$ can vary from engine to engine, and it is manufacturer-dependant. An adversary having this knowledge can perform the attacks more effectively. In a more practical sense, the degradation of the engine is very negligible at the beginning, but as time proceeds, the degradation follows a linear trend, and it increases as the engine approaches the end of life. Assuming in an engine, the linear degradation initially starts at $N^{d}$ cycle. The value of $N^{d}$ is different for different engines, as the wear of the engines may be different. If the average of $N^{d}$ for all the engines in the dataset is taken, it is found to be $N_{avg}^{d}$. An adversary knowing $N_{avg}^{d}$ can perform the attacks after the degradation initiates, making the attack more destructive. 

To study the impact of FDIA on PdM systems, we consider an attack scenario where the attacker has access to the aircraft and could attach a device to avionics CAN bus \cite{icsAlert} as mentioned previously in section IV (attack surface). The device attached to CAN bus can inject false data into engine sensor readings, resulting in incorrect predictions of RUL of the aircraft engine. Note, as mentioned in section (attack surface), it is also possible to launch an FDI attack without direct access to the aircraft by using the sensor spoofing technique~\cite{kerns2014unmanned}, or using a drone carrying a special device capable of interfering and impacting the on-board aircraft sensor measurements~\cite{mcas}. In this work, we consider two variations of FDIA which are continuous and interim FDIA. In continuous FDIA, the attack is initiated after $N^{d}$ and continues to the end-of-life of the engine. In Interim FDIA, the attack is initiated after $N^{d}$ and continues to the next 20 time cycles. In both the variations of FDIA, random and biased FDIAs are used to evaluate the PdM model's performance. Here, random FDIA means the noise added to the sensor output has a range (0.01\% to 0.05\%). Whereas, biased FDIA has a constant amount of noise added to the sensor output.

\begin{figure}[!h]
	\centering
	\includegraphics[width=0.35\textwidth]{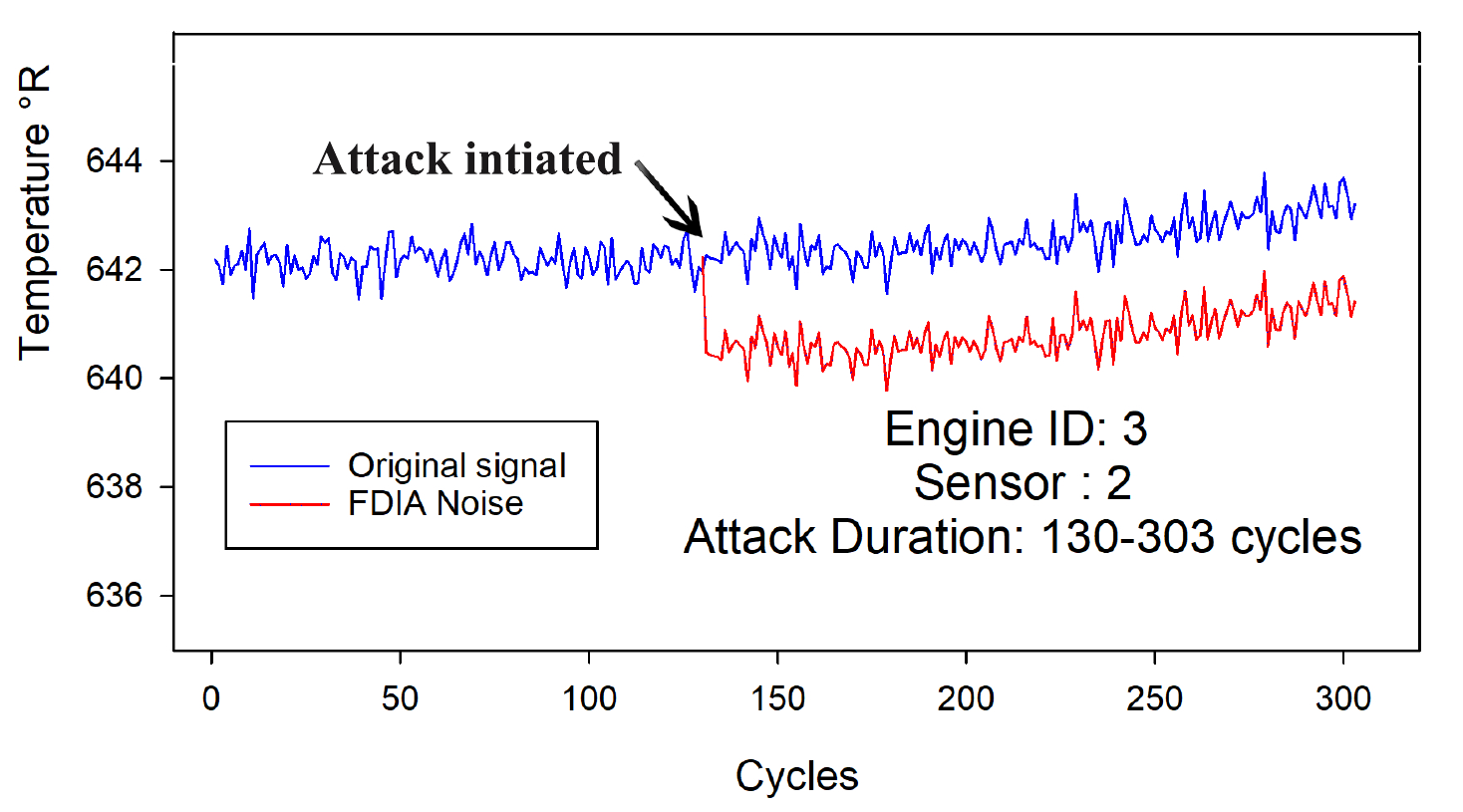}
	\caption{Continuous FDIA signature}
	\label{fig:sign1}
	\vspace{-3mm}
\end{figure}

\begin{figure}[!h]
	\centering
	\includegraphics[width=0.35\textwidth]{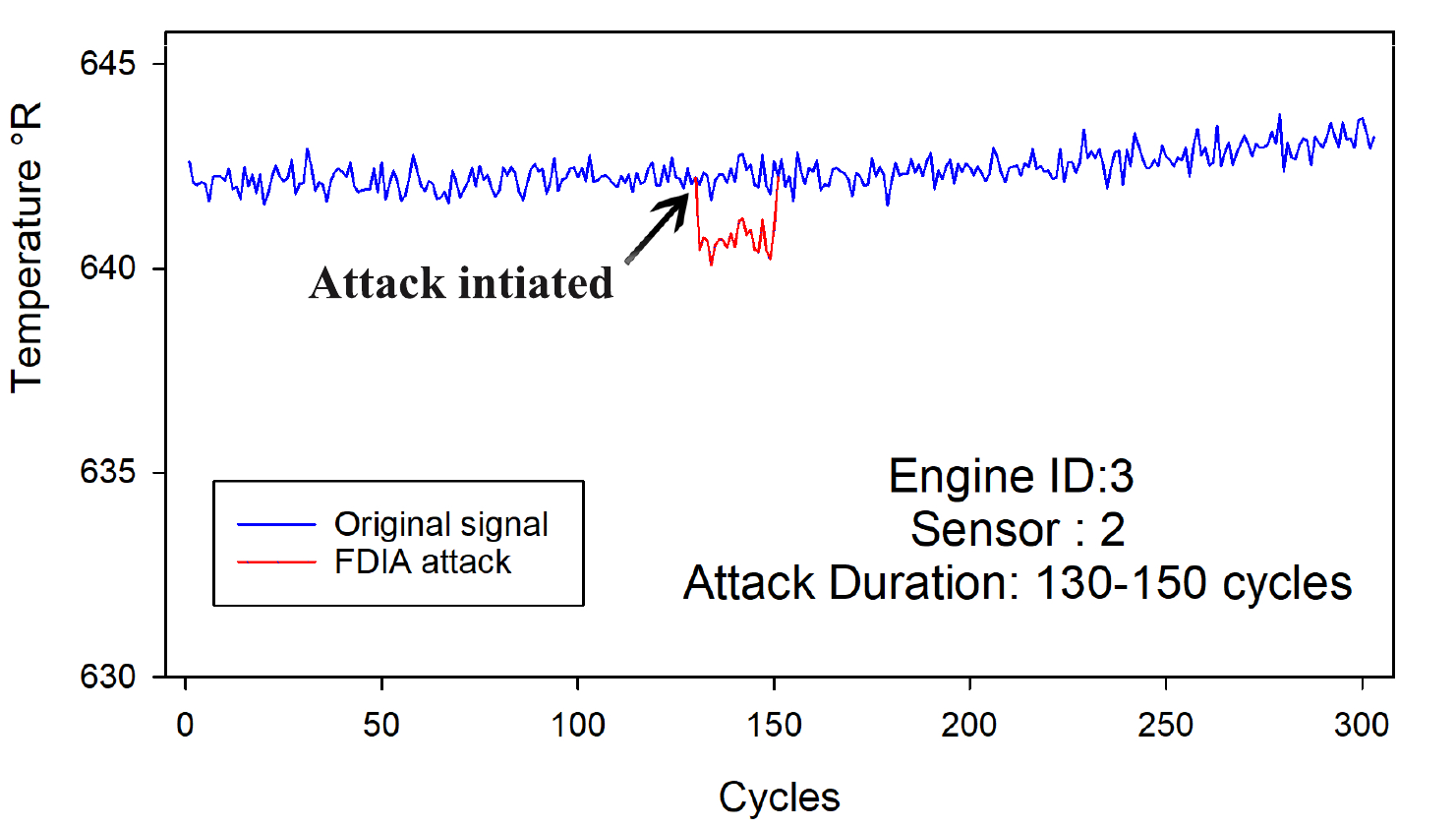}
	\caption{Interim FDIA signature}
	\label{fig:sign2}
\end{figure}

\begin{table}[h]
\centering
\caption {RMSE comparison for different DL algorithms}
\begin{tabular}{|c|c|}
\hline
\textbf{Predictor architecture}& \multicolumn{1}{c|}{\textbf{RMSE}} \\ \cline{2-2}
                              & \textbf{Test}                                \\ \hline
   CNN(64,64,64.64) lh(100)       & 9.94  \\ [0.5ex]                        
   LSTM(100,100,100,100) lh(80)    & 8.76 \\ [0.5ex] 
    GRU(100,100,100) lh(80)         & 7.26\\ [0.5ex]
 \hline
\end{tabular}

\label{tab:MEATable}
\end{table}

\begin{figure*}[ht]
	\centering
	\includegraphics[width=0.9\textwidth]{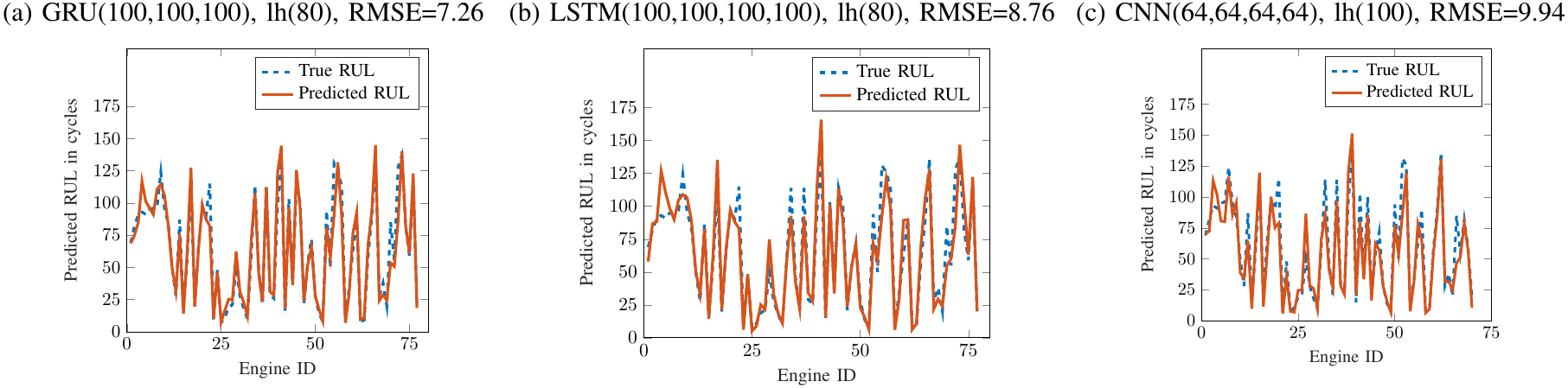}
	\caption{Comparison of deep learning algorithms}
	\label{fig:LSTMStruct}
	\vspace{-2mm}
\end{figure*}
\section{EXPERIMENTAL RESULTS} \label{sec:Exp}
In this section, we first compare different DL algorithms on RUL prediction. Next, we present both continuous and interim FDIA signatures, and the impact of attacks on the RUL prediction. Lastly, we present piece-wise RUL prediction and detail the impact of sequence length on resiliency. For the sake of reproducibility and to allow the research community to build on our findings, the artifacts (source code, datasets, etc.) of the following experiments are publicly available on our GitHub repository\footnote{https://github.com/dependable-cps/FDIA-PdM}.

\subsection{Comparison of deep learning algorithms} \label{sec:Comp}
In order to select the best machine learning algorithm for the PdM, we compare LSTM, GRU, and CNN algorithms for the C-MAPSS dataset. To evaluate the performance of the predictors, we utilize the root mean square error (RMSE) metric which is widely used as an evaluation metric in model evaluation studies. Fig. \ref{fig:LSTMStruct} and Table \ref{tab:MEATable} represents the comparison of DL algorithms with architectures LSTM(100,100,100,100) lh(80), GRU(100,100,100) lh(80), and CNN(64,64,64,64) lh(100). The notation GRU(100,100,100) lh(80) refers to a network that has 100 nodes in the hidden layers of the first GRU layer, 100 nodes in the hidden layers of the second GRU layer, 100 nodes in the hidden layers of the third GRU layer, and a sequence length of 80. In the end, there is a 1-dimensional output layer. Table \ref{tab:Para} shows the hyperparameters for the developed CNN, LSTM and GRU models (inspired from \cite{ellefsen2019remaining} and \cite{ince2016real}, with additional experiments to check the feasibility of the adopted hyperparameters).



\begin{table}[h]
\centering
\caption{Hyperparameter settings}
\begin{tabularx}{0.45\textwidth}{|c|XXXXX|} 
\hline 
	\textbf{Model}             & 	\textbf{Hidden neuron} & 	\textbf{Dropout} & 	\textbf{Batch size} & 	\textbf{Epochs} &  \textbf{Act. func.}\\ [0.5ex] 
\hline
CNN     & 64    & 0.2   & 200   & 100   & ReLu \\ [0.5ex] 
LSTM    & 100   & 0.2   & 200   & 100   & tanh \\ [0.5ex] 
GRU     & 100   & 0.2   & 200   & 100   & tanh \\ [0.5ex] 
\hline
\end{tabularx}
\label{tab:Para}  

\end{table}



From Fig. \ref{fig:LSTMStruct} and Table \ref{tab:MEATable} it is evident that the DL algorithm GRU(100, 100, 100) with a sequence length 80 has the least RMSE of 7.26. It means that GRU is very accurate in predicting accurate RUL for this dataset. Note, the obtained results in Table \ref{tab:MEATable} show that our GRU-based predictive model performs 1.9, 1.7 and 1.3 times better (in terms of accuracy) when compared to the recent works in \cite{zheng2018data,ellefsen2019remaining,wang2019deep}, respectively, on RUL estimation using DL algorithms and the C-MAPSS dataset. In the next step, we model FDIA on CNN, LSTM, and GRU to evaluate their resiliency to the FDIA.




\begin{figure*}[t]
	\centering
	\includegraphics[width=0.8\textwidth]{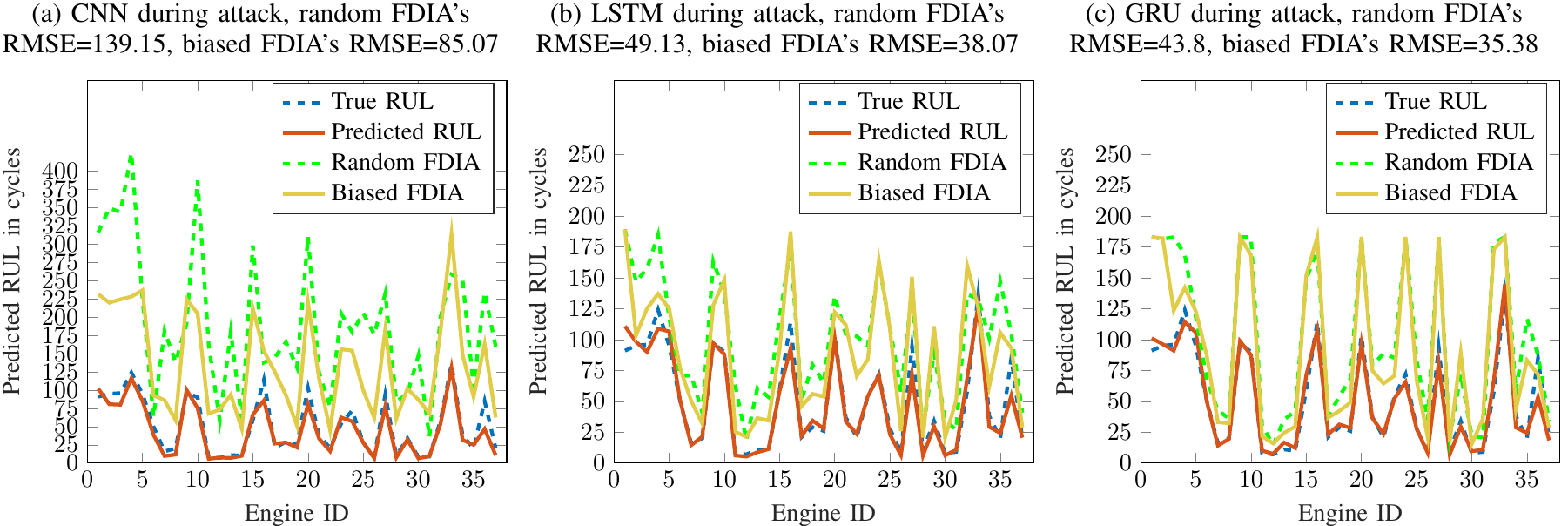}
	\caption{FDI attack scenario for continuous period}
	\label{fig:Scen1}
\end{figure*}


\begin{figure*}[t]
	\centering
	\includegraphics[width=0.8\textwidth]{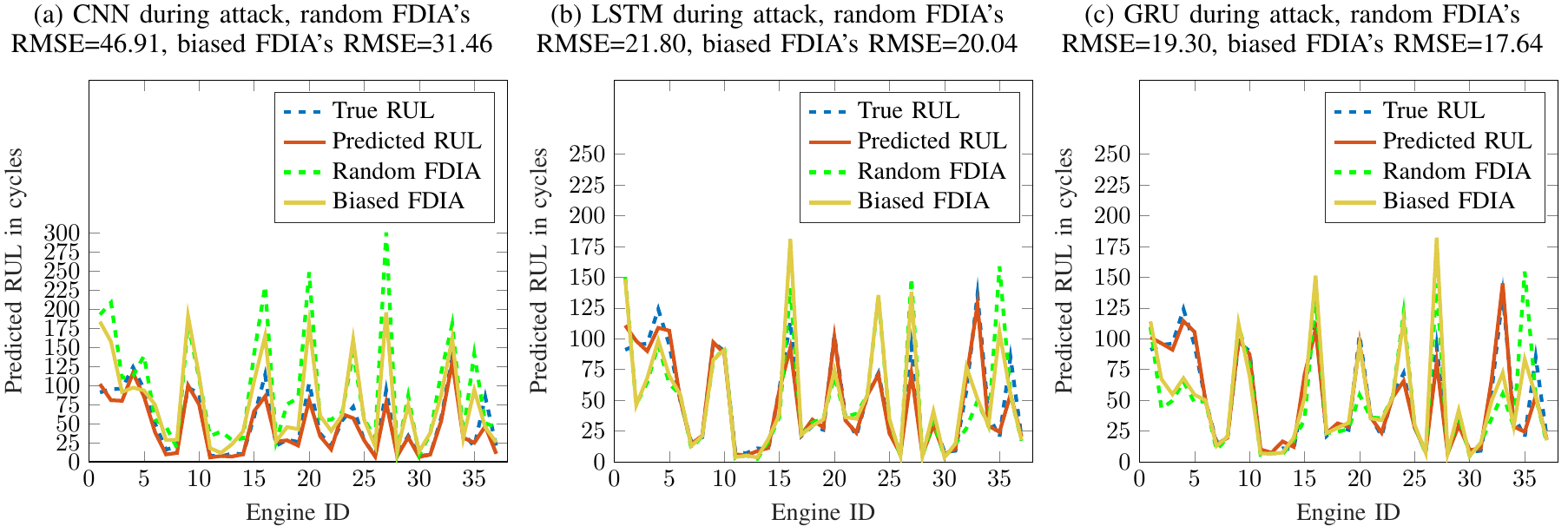}
	\caption{FDI attack scenario for interim period}
	\label{fig:Scen2}
\end{figure*}


\begin{figure*}[t]
	\centering
	\includegraphics[width=0.8\textwidth]{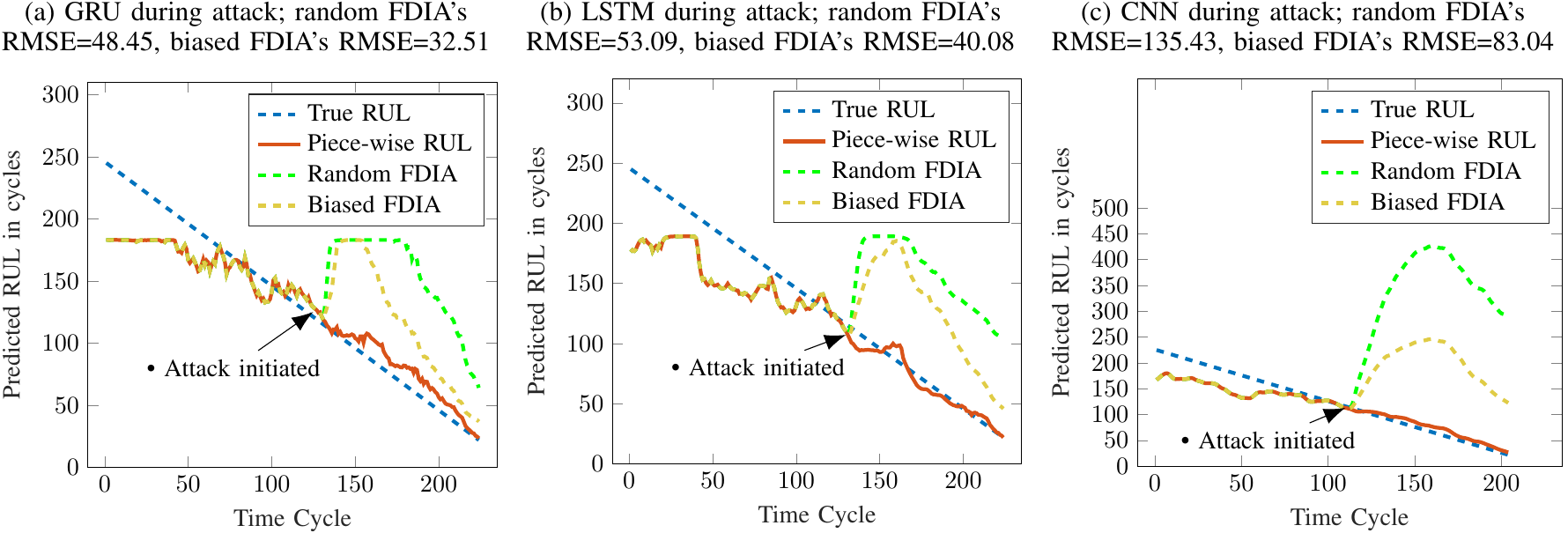}
	\caption{Piece-wise RUL prediction for continuous FDIA}
	\label{fig:piecewise1}
	\vspace{-2mm}
\end{figure*}


\begin{figure*}[t]
	\centering
	\includegraphics[width=0.8\textwidth]{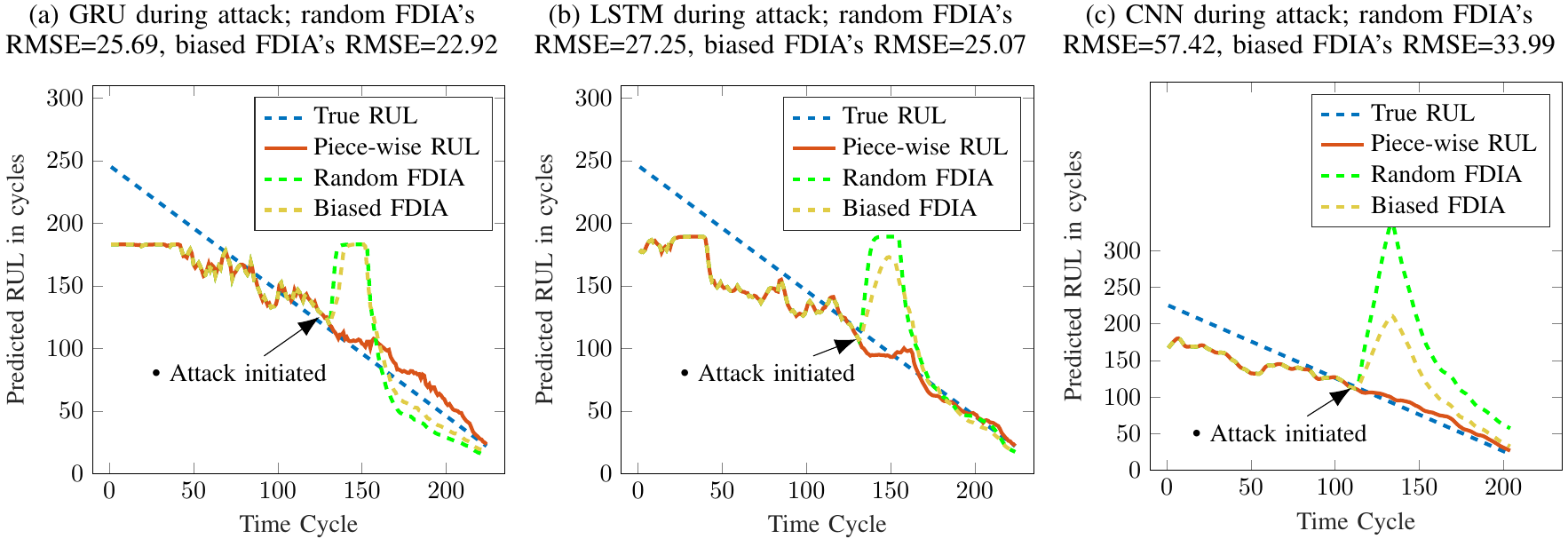}
	\caption{Piece-wise RUL prediction for Interim FDIA}
	\label{fig:piecewise2}
	\vspace{-2mm}
\end{figure*}

\subsection{Impact of attacks on a PdM system} \label{sec:Cyber}
The average degradation point of the engine $N_{avg}^{d}$ is considered as 130 for the FD001 dataset \cite{130a}~\cite{130b}~\cite{main}, and we assume that the Engine Health Monitoring (EHM) system of the aircraft sends 20-time cycles ($N_b$) of data to the ground at a time as shown in Fig. \ref{fig:EHM}. The train and test dataset have 21 sensor data. The FDIA can be performed on 21 sensors, but to make the attack more realistic, we perform FDIA on only 3 sensors (specifically, T24, T50, and P30). Details about these 21 sensors can be found in \cite{sensorsTable}. In FDIA continuous scenario, the attacker has initiated the attacks after $N_{avg}^{d}$, which is 130-time cycles (a one-time cycle is equivalent of one flight hour), and the attack duration is until end of life of the engine. In FDIA interim scenario, the attacker has initiated the attacks after $N_{avg}^{d}$, which is 130-time cycles, and the attack duration is 20 hours (20-time cycles). Since the attack is initiated after 130-time cycles, we only consider the engines which have data for more than 130 cycles which gives us 37 engines in the FD001 dataset. The resultant dataset is re-evaluated using the LSTM, CNN and GRU-based PdM models and the obtained RMSEs are 6.09, 7.50, and 5.36, respectively.\\

\noindent \textbf{FDIA signature:}
To model the FDIA on sensors, we add a vicious vector to the original vector, which modifies the sensor output by a very small margin (0.01\% to 0.05\%) for random FDIA and 0.02\% for biased FDIA. Here, random FDIA means the noise added to the sensor output has a range (0.01\% to 0.05\%). Whereas, biased FDIA has a constant amount of noise added to the sensor output. Fig. \ref{fig:sign1} shows the comparison between the original and biased FDIA attacked output signal of sensor 2 for engine ID 3 for a continuous period. In continuous FDIA, we attack the sensor output from time cycles 130 to the end-of-life of the engine. In the case of biased FDIA for an interim period as shown in Fig. \ref{fig:sign2}, the attack duration is only for 20 time cycles (130 to 150 time cycles). Note, in the constrained attack the adversary has limited access to sensors. As shown in Fig. \ref{fig:sign1} and \ref{fig:sign2}, the attack signature is very similar to the original signal, making it stealthy and harder to detect even with common defense mechanisms in place.\\
    
	
\noindent \textbf{Impact of FDIA on CNN, LSTM and GRU:}
To show the impact of an FDIA on the aircraft PdM system, we implement an attack for the scenario mentioned previously in Section IV (attack scenario). The FDIA is performed on three sensors (T24, T50, and P30) instead of attacking all the 21 sensors in the dataset. In FDIA continuous scenario, the adversary performs attacks from 130-time cycles to end of life of the engine. It is evident from Fig. \ref{fig:Scen1} that LSTM, GRU, and CNN are greatly affected by the continuous FDI attack. In the case of random and biased FDIA, random FDIA showed a considerable impact on all PdM models. The CNN based PdM model is the most affected by the continuous FDIA as random FDIA's RMSE is 139.15 and biased FDIA's RMSE is 85.07 (true RMSE is 7.50) which is almost 18 times and 11 times higher when compared to the true RMSE, respectively. In contrast, the GRU based PdM model is the least affected by the continuous FDIA as random FDIA's RMSE is 43.8 and biased FDIA's RMSE is 35.38 (true RMSE is 5.36). Even though the GRU is least affected by both random and biased FDIA, their RMSE is 8 and 6 times higher than the true RMSE, respectively, making it also deadly for a PdM system.

In the FDIA interim scenario, the adversary performs attacks between 130 and 150-time cycles (20-time cycles). It is evident from Fig. \ref{fig:Scen2} that LSTM, GRU, and CNN are greatly affected by the interim FDI attack. Once again, the CNN based PdM model is greatly affected by the continuous FDIA as random FDIA's RMSE is 46.91 and biased FDIA's RMSE is 31.46 (true RMSE is 7.50) which is almost 6 times and 4 times higher than the true RMSE, respectively. In contrast, the GRU based PdM model is the least affected by the interim FDIA as random FDIA's RMSE is 19.30 and biased FDIA's RMSE is 17.64 (true RMSE is 5.36). This indicates that GRU-based PdM models are comparatively resilient to both continuous and interim FDIA. Even though the GRU is least affected by both random and biased FDIA, their RMSE is still 4 times and 3 times higher than the true RMSE, respectively, making it deadly for a PdM system. When comparing both continuous and interim FDIA, it observed that continuous FDIA's RMSE is almost twice the interim FDIA's RMSE. Hence, continuous FDIAs are more potent than interim FDIA.




\subsection{Piece-wise RUL prediction} \label{sec:Piece}

In order to show the impact of FDIA attacks on a specific engine data, we apply the piece-wise RUL prediction. The piece-wise RUL prediction gives a better visual representation of degradation in an aircraft engine. Fig. \ref{fig:piecewise1}(a) shows an example of an engine data from the dataset of 100 engines, and depicts the predicted RUL using GRU at each time step of that engine data. For example, if $X$ is the time series data of a particular engine, then  $X_i=[x_1,  x_2,  x_3...  x_{t-k}]$ represents time series data until time $t-k$. $RUL^{p}$ is predicted RUL at each time step in $X$, which is can be defined as $RUL_i^{p}=[RUL_1^{p},  RUL_2^{p},  RUL_3^{p}... RUL_{t-k}^{p}]$. From Fig. \ref{fig:piecewise1}(a), it is evident that as the time series approaches the end of life, the predicted RUL (red line) is close to the true RUL (blue dashes), because the DL model has more time series data to accurately predict the RUL. 


In the case of piece-wise RUL prediction during continuous FDIA, it is observed from Fig. \ref{fig:piecewise1} that both random and biased FDIAs are initiated from 130-time cycles to 242-time cycles for engine ID 17. Here, the green and yellow dashes in the figures are predicted RUL after random and biased FDIA, respectively. In the GRU, LSTM, and CNN based piece-wise RUL prediction (for both random and biased FDIA), the attacker initiates the FDIA after 130-time cycles. The impact of the attack is quite interesting as the RUL jumps upwards (around 200 for GRU and LSTM) with a possible indication to the engine maintenance operator that the engine is quite healthy. This may influence a `no maintenance required' decision from the maintenance engineers' point of view, however, in reality, the RUL is decreasing continuously and going below the 100-time cycles which might require to schedule urgent maintenance leading to a catastrophic event. For CNN, the continuous FDIA causes a longer jump (even beyond the initial RUL value) when compared to the FDIA in LSTM and GRU. Of course, there is a higher chance that this will be flagged as a potential fault either in the engine or in the PdM system, and will cause unnecessary engine maintenance and will increase the aircraft downtown causing a financial loss to the flight operator. 

In the case of piece-wise RUL prediction for engine ID 17 under interim FDIA, it is observed in Fig. \ref{fig:piecewise2} that the attack causes a similar jump as shown in the case of continuous FDIA in Fig. \ref{fig:piecewise1}. However, the effect of the attack flushes away way sooner when compared to the continuous FDIA case. However, note that the attack duration was only 20 cycles, but it took more than 45 cycles to flush out the effect by the PdM system. Hence, if maintenance is due around that period, it may lead to catastrophic consequences. Once again, the piece-wise RUL prediction results indicate that employing CNN in PdM systems may result in systems that are very sensitive to the FDIA and hence special measures should be taken for designing a CNN-based PdM.  
\subsection{Impact of sequence length on resiliency of GRU} \label{sec:SeqLen}
\begin{figure}[h]
	\includegraphics[width=0.49\textwidth]{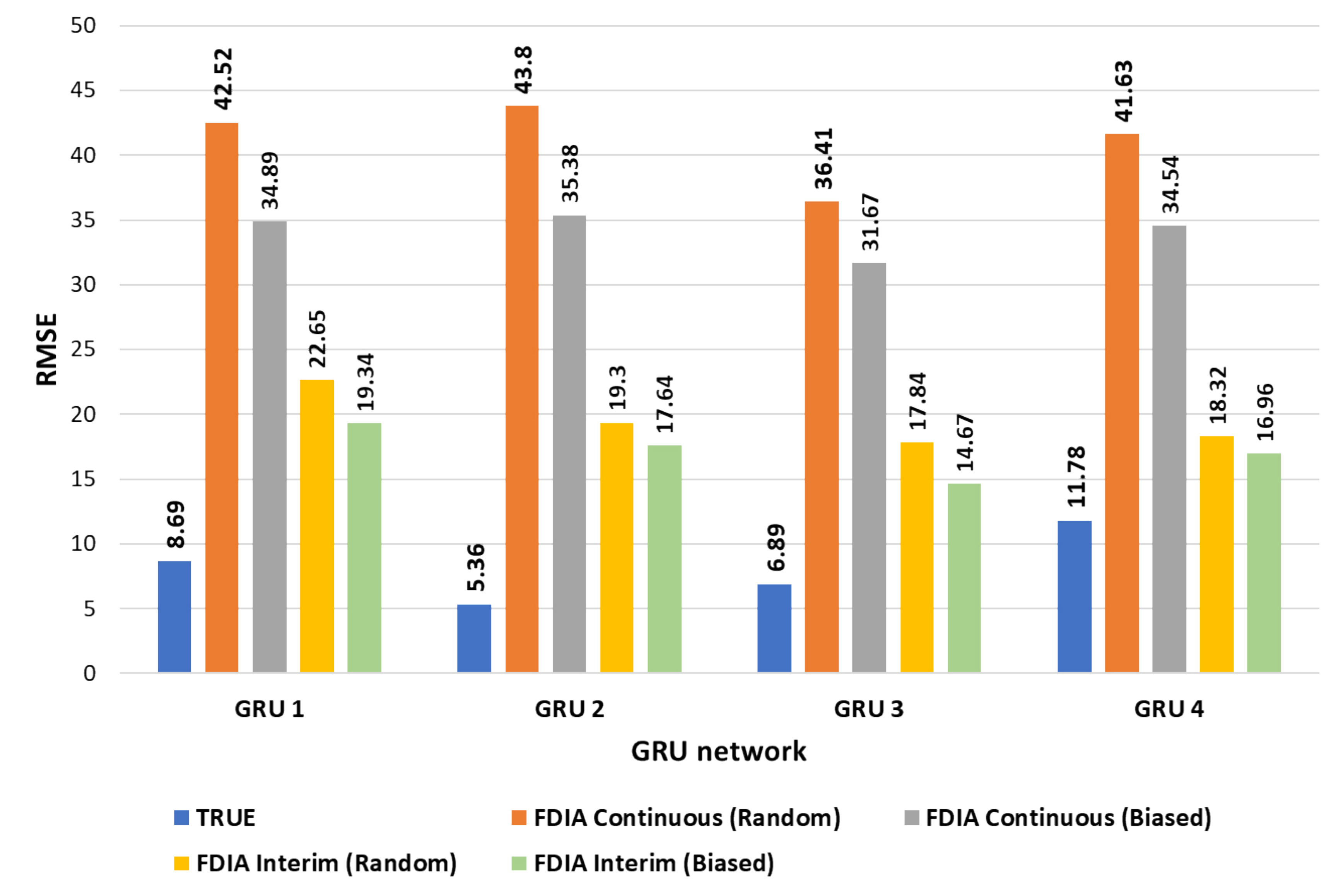}
	\caption{RMSE comparison of different GRU networks}
	\label{fig:GRUComp}
	\vspace{-2mm}
\end{figure}

Since GRU has performed best among the DL algorithms as shown in the experimental results in the previous subsections, in \figurename~\ref{fig:GRUComp} we compare four different GRU networks under FDI attack. The GRU networks have structures GRU1(100,100,100) lh(90), GRU2(100,100,100) lh(80), GRU3(100,100,100) lh(70), and GRU4(100,100,100) lh(60). We observe that the GRU network with architecture GRU2(100,100,100) lh 80 has the least value of true RMSE (5.36), which means that it predicts RUL quite accurately, however, it is less resilient to both continuous and interim FDIA. In contrast, GRU with network architecture GRU3(100,100,100) lh(70) shows the second-best performance in predicting the RUL (RMSE of 6.89), however, in terms of resiliency, this network is the least affected by continuous and interim FDIA. This indeed shows an interesting insight that the sequence length affects not only the accuracy but also the resiliency of the model. It also indicates that accuracy should not be the only factor while designing a PdM system. For instance, in terms of accuracy GRU2 is the typical choice. However, if both accuracy and resiliency are considered, GRU3 is can be an ideal choice (at the cost of losing some accuracy).

\section{Discussion} \label{sec:Obs}
In this work, we first evaluate different deep learning (DL) algorithms on the C-MAPSS dataset and obtained results show a great prospect for deep learning in PdM. It is also observed that sequence length and network architecture are crucial in predicting accurate RUL. Our work shows that the GRU performed 1.3-1.9 times better than the recent works that use deep learning on the C-MAPSS dataset \cite{zheng2018data,ellefsen2019remaining,wang2019deep}.


The impact analysis of FDIA on aircraft sensors in the C-MAPSS dataset provides some interesting insights. We observe that CNN based PdM model is greatly affected by both random and biased FDIA. In the case of interim FDIA, CNN's random and biased RMSE are 18 and 11 times higher than the true RMSE, respectively, and in the case of continuous, the random and biased RMSE are 6 and 4 times higher than the true RMSE, respectively. We also observe that the GRU-based PdM model is more resilient to both random and biased in comparison with CNN and LSTM-based PdM models. Even though the GRU is least affected by both random and biased FDIA, their RMSE is 8 and 6 times higher than the true RMSE in the case of continuous FDIA, respectively. In the case of interim FDIA, the random and biased RMSE are 4 and 3 times higher than the true RMSE, respectively, making it disastrous for the PdM system. This may result in the delay of timely maintenance for the aircraft engine and eventually result in engine failure at some point. Note, the attack signature of FDIA is very close to the original sensor output (within the boundary conditions of the sensor measurements) making it harder to be detected by common defense mechanisms in an engine health monitoring (EHM) system. 


A piece-wise RUL predicting approach is used in visualizing the impact of attacks on the sensors, which clearly shows that the PdM system is susceptible to sensor attacks. While designing of PdM systems, the engineers should take both continuous and interim FDI attacks into consideration. CNN based piece-wise RUL prediction results show that special measures should be taken when designing and adopting CNN-based PdM systems (such as the cases in~\cite{silva2019cnn,gunnemann2017predicting,huuhtanen2018predictive,caponetto2019deep}) as they are very sensitive to the FDIA. \figurename \ref{fig:GRUComp}, gives an interesting insight into the relationship between accuracy and resiliency of the GRU network. It shows the need for considering the relationship between the accuracy, resiliency and sequence length of a DL mode (such as GRU in our case) in the design phase. Indeed, such an analysis can serve as empirical guidance to the development of subsequent data-driven PdM systems.

All of these obtained results show that DL-based PdM systems have a great prospect for aircraft maintenance, however, they are very susceptible to sensor attacks. Hence it is required to investigate proper detection techniques to detect such stealthy attacks and special care should be taken when manufacturing IoT sensors for DL/AI applications. For the same reason, while designing a PdM system, the designer also must consider the resiliency of the DL algorithm instead of just emphasizing on the algorithm's accuracy, as we investigated in this paper. 


\section{Conclusions and Future Works}
\label{Sec:Conc}
This paper compares the performance of LSTM, GRU, and CNN for RUL prediction using the C-MAPSS dataset, and explores the impacts of continuous and interim FDI attacks on these deep learning algorithms. We observe that the GRU is a better suited DL technique when compared to LSTM and CNN in terms of accuracy. The obtained results show that both continuous and interim FDIA have a substantial impact on the RUL prediction even if only a few IoT sensors are attacked. We also observed that the GRU-based PdM model is more resilient to FDIA, whereas CNN is dramatically sensitive to both continuous and interim FDIA. Finally, we explored that there exists a relationship between the accuracy and sequence length in the GRU-based PdM model which can serve as empirical guidance to the development of data-driven PdM systems. In the future, we plan to develop an end-to-end methodology for the detection and mitigation of sensor attacks in a PdM system. 

\bibliographystyle{IEEEtran}
\bibliography{ref}

\end{document}